\documentclass[10pt,twocolumn,letterpaper]{article}

\usepackage{iccv}
\usepackage{times}
\usepackage{epsfig}
\usepackage{graphicx}
\usepackage{amsmath}
\usepackage{amssymb}
\usepackage{multirow}
\usepackage{booktabs}
\usepackage{algorithm}
\usepackage{algorithmic}
\usepackage{comment}
\usepackage{subcaption}
\usepackage{xcolor}
\usepackage{enumitem}

\usepackage[pagebackref=true,breaklinks=true,letterpaper=true,colorlinks,bookmarks=false]{hyperref}

\iccvfinalcopy % *** Uncomment this line for the final submission

\def\iccvPaperID{XXXX} % *** Enter the ICCV Paper ID here

\ificcvfinal\fi

\begin{document}

%%%%%%%%% TITLE
\title{Learning Feature-to-Feature Translator \\ by Alternating Back-Propagation for Generative Zero-Shot Learning}

\author{Yizhe Zhu$^{1}$\thanks{Work was done while Yizhe Zhu was an intern at Hikvision.}, \quad Jianwen Xie$^{2}$, \quad  Bingchen Liu$^{1}$, \quad Ahmed Elgammal$^{1}$\\
$^{1}$Department of Computer Science, Rutgers University \quad $^{2}$Hikvision Research Institute \\
{\tt\small yizhe.zhu@rutgers.edu, jianwen@ucla.edu, bingchen.liu@rutgers.edu,  elgammal@cs.rutgers.edu }
}

\maketitle

%%%%%%%%% BODY TEXT
\begin{abstract}
We investigate learning feature-to-feature translator networks by alternating back-propagation as a general-purpose solution to zero-shot learning (ZSL) problems. It is a generative model-based ZSL framework. In contrast to models based on generative adversarial networks (GAN) or variational autoencoders (VAE) that require auxiliary networks to assist the training, our model consists of a single conditional generator that maps class-level semantic features and Gaussian white noise vector accounting for instance-level latent factors to visual features, and is trained by maximum likelihood estimation. The training process is a simple yet effective alternating back-propagation process that iterates the following two steps: (i) the inferential back-propagation to infer the latent factors of each observed example, and (ii) the learning back-propagation to update the model parameters. We show that, with slight modifications, our model is capable of learning from incomplete visual features for ZSL. We conduct extensive comparisons with existing generative ZSL methods on five benchmarks, demonstrating the superiority of our method in not only ZSL performance but also convergence speed and computational cost. Specifically, our model outperforms the existing state-of-the-art methods by a remarkable margin up to $3.1\%$  and $4.0\%$ in ZSL and generalized ZSL settings, respectively. 
\end{abstract}

\section{Introduction}
Deep learning techniques have successfully tackled various computer vision problems such as object detection~\cite{he2017mask,ren2015faster,fan2019shifting,liu2016ssd,fan2018salient,zhao2019contrast}, image, video and 3d shape generation~\cite{kingma2013auto, goodfellow2014generative,  xie2016cooperative,xie2016theory,reed2016generative, xie2017synthesizing, xie2018learning, zhao2018learning, tian2018cr}, pose estimation~\cite{peng2018jointly,tang2019adatransform,zhao2019semantic,tang2018quantized,tang2018cu}, object recognition~\cite{krizhevsky2012imagenet, szegedy2015going, he2016deep}, etc. Especially, these advanced deep learning methods equip the machine with the comparable ability of object recognition to human beings when abundant labeled training samples are provided. However, this ability will decrease dramatically for classes that are insufficiently represented or even not present in the training data, thus increasing the difficulty of real-world applications due to the costly data collection and annotation effort. This limitation attracts the intense interest of researchers in zero-shot learning (ZSL). 

\begin{figure}[t]
	\centering
	\includegraphics[width=1.0\columnwidth]{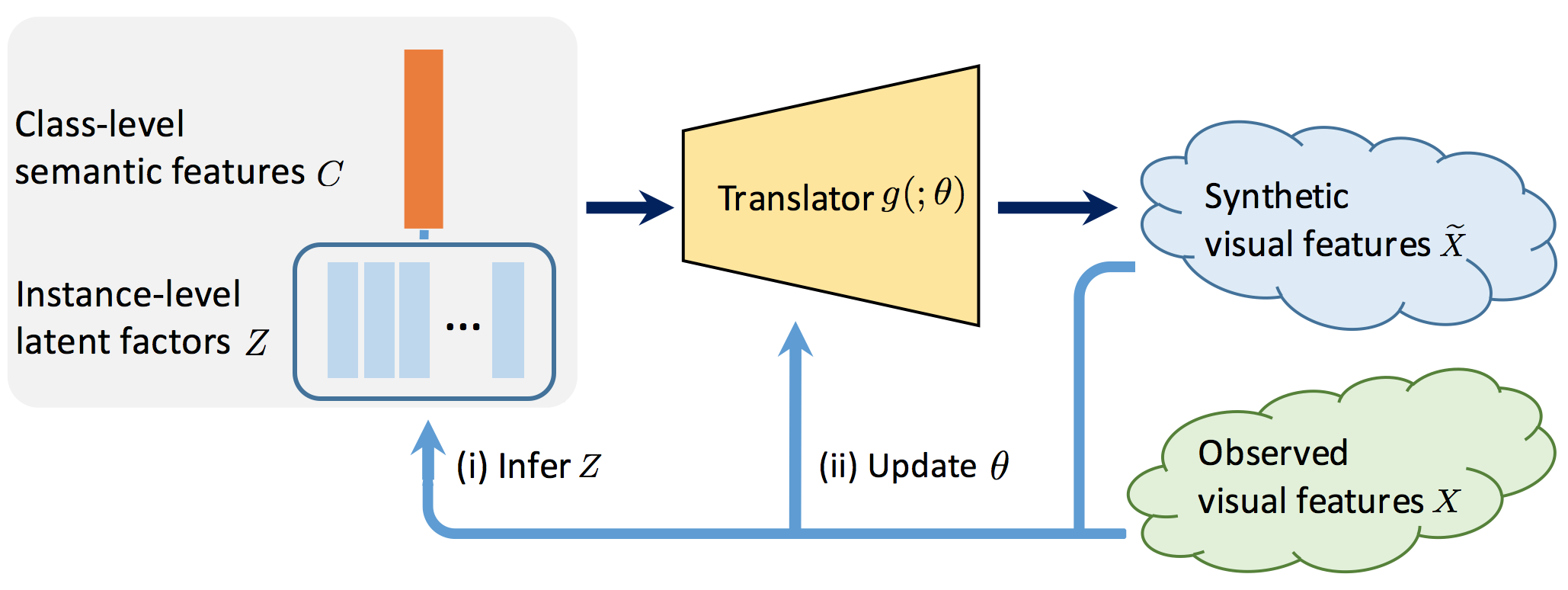}	
	\caption{Demonstration of the feature-to-feature translator for generative zero-shot learning. Training (blue flow): given the class-level semantic features $C$ and the observed visual features $X$, (i) the latent factors $Z$ are inferred by MCMC and (ii) the weights $\theta$ of the translator are updated by the gradient ascent for maximum likelihood. Translation (black flow): once the translator is learned, arbitrary number of synthetic visual features can be generated for unseen classes by translating the unseen class semantic features along with $Z$ randomly sampled from Gaussian distribution. The generated samples for unseen classes are useful for zero-shot learning.}
	\label{fig:intro}
	\vspace{-2em}
\end{figure}

ZSL aims to recognize novel classes where no training data is available for these classes. The key to make ZSL work is to use the semantic description of classes as the bridge to connect the seen classes and unseen classes. 

Recently, generative ZSL approaches ~\cite{bucher2017generating, zhu2018generative,xian2018feature,felix2018multi, wang2017zero, verma2018generalized} have emerged as a new trend of ZSL strategy by exploiting the successful generative models, e.g., variational autoencoders (VAE) ~\cite{kingma2013auto, rezende2014stochastic}, generative adversarial networks (GAN)~\cite{goodfellow2014generative, mirza2014conditional}, etc, to learn mappings  from class-level semantic features (e.g., attributes or description embeddings) to visual features.  The synthetic visual features for unseen classes through well-trained generative models establish the visual description of unseen classes and make the conventional supervised classification applicable.

The quality of generative ZSL approaches mainly depends on how well the generative model can emulate the data distribution. GAN-based methods~\cite{zhu2018generative,xian2018feature,felix2018multi} employ the conditional generator network which maps class-level semantic features along with Gaussian white noise as latent factors to visual features. The conditional generator is trained in an adversarial manner where a well-designed discriminator network is recruited to play a minimax game with the conditional generator. The adversarial training strategy gets around the intractable inference of latent factors in training; however, the imbalance between the discriminator and the generator would lead to non-convergence and mode collapse issues. VAE-based methods~\cite{wang2017zero,verma2018generalized} associate the conditional generator network with an additional encoder that approximates the posterior distribution for the purpose of inference of latent factors, and train both models by maximizing the variational lower bound. In this paper, we will show that neither of these assisting networks is necessary for training the conditional generator. 

In our framework of generative ZSL, we adopt a conditional generator as a feature-to-feature translator network that translates the class-level semantic features along with the instance-level latent factors to visual features as shown in Figure~\ref{fig:intro}. However, different from GAN and VAE-based methods, we resort to a theoretically more accurate estimator, maximum likelihood estimation (MLE) with EM-like strategy~\cite{dempster1977maximum}, to train the network, without the requirement of auxiliary networks for assistance. 
The hard nut to crack in training latent variable models by MLE is the intractable posterior distribution that is required for computing the gradient of the observed-data log-likelihood. In the proposed framework, we adopt Markov chain Monte Carlo (MCMC), such as Langevin dynamics ~\cite{welling2011bayesian, neal2011mcmc}, for computing the posterior distribution. We show that the maximum likelihood algorithm involves the computation of the gradient of observed-data log-likelihood with respect to model parameters and the gradient with respect to latent factors, both of which can be efficiently computed by back-propagation. 

Specifically, the proposed translator is learned by alternating back-propagation (ABP) algorithm, which iterates the following two steps: (i) \emph{inferential back-propagation}: for each training example, inferring the continuous latent factors by sampling from the current learned posterior distribution via Langevin dynamics, where the gradient of the log joint density can be calculated by back-propagation, (ii) \emph{learning back-propagation}: updating the model parameters given the inferred latent factors and the training examples by gradient ascent, where the gradient of the log-likelihood with respect to the model parameters can again be calculated by back-propagation. 

The ABP algorithm was originally proposed for training unconditional generator networks \cite{han2017alternating, xie2019learning}. In this paper, we generalize ABP to learning conditional generator network for feature-to-feature translation, where the class-level semantic features play the role of condition.

It is worth mentioning that the proposed model is capable of learning from incomplete training examples, where visual features are partially corrupted or inaccessible due to occlusion in image space. Specifically, our model can learn from incomplete data by making latent factors only explain the visible parts of the visual features conditioned on class-level semantic features, while GAN or VAE-based methods can hardly deal with this situation. 

Our contributions can be summarized as follows:
(1) We propose a feature-to-feature translator for generative ZSL, where we learn a mapping from class-level semantic features, along with instance-level latent factors following Gaussian white noise distribution, to visual features. (2) We propose to learn the translator via alternating back-propagation (ABP) algorithm for maximum likelihood, without relying on other assisting networks, which makes our framework more statistically rigorous and elegant. (3) We show that the proposed framework can learn from incomplete training examples where visual features are partially visible due to corruption or occlusion in image space. (4) Comprehensive experiments conducted on various ZSL tasks show the state-of-the-art performance of our framework, and a thorough analysis of the model demonstrates the superiority of the proposed framework in different aspects. 

\section{Related Work}
\textbf{X-to-Y Translation}.  We are not the first to apply conditional generators to learn X-to-Y mapping. A text-to-image translation is proposed for image synthesis from text description~\cite{reed2016generative}. Zhu \emph{et al}~\cite{zhu2017unpaired} has studied image-to-image translation problem for different types of image processing tasks, which include synthesizing photos from label maps or edge maps and generating color images from their grey-scaled versions. Recently, video-to-video translation problem has been tackled by learning a mapping function from an input source video (e.g., a sequence of semantic segmentation masks) to a target realistic video~\cite{wang2018vid2vid}. Our work learns a feature-to-feature mapping for ZSL.
Additionally, all other works mentioned above are based on the framework of GANs, which means that well-designed discriminator networks need to be resorted to in the training stage. Our framework differs in that it is trained by an alternating back-propagation algorithm without incorporating any extra assisting networks. This makes our framework considerably simpler and computationally more efficient than those based on GANs.

\textbf{Generative Models}. Our model is essentially a conditional latent variable model. The alternating back-propagation (ABP) training method for our model is related to variational inference (e.g., VAE) and adversarial learning (e.g., GAN), both of which require an extra assisting network with a separate set of learning parameters to avoid the explaining-away inference of latent variables in the model. 

Unlike VAE and GAN, ABP does not involve an auxiliary network and performs explicit explaining-away inference by directly sampling from the posterior distribution via MCMC, such as Langevin dynamics, which is powered by back-propagation. Our model trained by ABP is much simpler, more natural and statistically rigorous than those trained by adversarial learning and variational inference schemes. 
ABP has been used to train general generators~\cite{han2017alternating} and deformable generators \cite{xing2018deformable} for image patterns, as well as dynamic generators~\cite{xie2019learning} for video patterns. Our paper is a generalization of \cite{han2017alternating, xie2019learning} by applying ABP to train a conditional version of the generator model for feature-to-feature translation. The generative ConvNet \cite{xie2016theory} and the Wasserstein INN \cite{lee2018wasserstein} are two one-piece models that learn energy-based generative models for data generation. Both \cite{xie2016theory} and \cite{lee2018wasserstein} generate data via iterative MCMC sampling, while our model generates data via direct ancestral sampling, which is much more efficient. 

\textbf{Zero-Shot Learning}. Several pioneering works for ZSL ~\cite{lampert2009learning, lampert2014attribute} make use of class attributes as intermediate information to classifiy images for unseen classes. Some ZSL methods are based on the bilinear compatibility function between the visual and semantic features, which can be learned by using (a) the ranking loss (e.g., ALE~\cite{akata2016label} and DeViSE~\cite{frome2013devise}), (b) the structural SVM loss ~\cite{akata2015evaluation} or (c) the ridge regression loss (e.g., ESZSL~\cite{romera2015embarrassingly} and PTZSL~\cite{Elhosein_2017_CVPR}). To enhance the expressive power of the models, several ZSL approaches~\cite{xian2016latent,zhang2017learning,li2019rethinking,li2019on,zhu2019learning} learn non-linear multimodal embedding.
Taking advantage of the generative models in data generation, several methods~\cite{zhu2018generative,felix2018multi, wen2016discriminative, verma2018generalized, bucher2017generating,wang2017zero} resort to generating visual features from unseen classes for ZSL. Both GAZSL~\cite{zhu2018generative} and FGZSL~\cite{wen2016discriminative}
pair a Wasserstein GAN~\cite{arjovsky2017wasserstein,gulrajani2017improved} with a classification loss as regularization to increase the inter-class  discrimination of synthetic features. MCGZSL~\cite{felix2018multi}
adopts cycle consistency loss ~\cite{zhu2017unpaired} to regularize the generator for ZSL . \cite{wang2017zero, verma2018generalized} employ conditional VAEs~\cite{sohn2015learning} framework to learn feature generator. Our model also learns to generate visual features from unseen classes by a conditional generator, however, different from the generative ZSL methods mentioned above, our model is trained by alternating back-propagation, without the need of assisting models for training. 

\section{Feature-to-Feature Translator}\label{sec:learning}
\subsection{Conditional Latent Variable Model for Feature-to-Feature Translation} 
Let $S=\{(X_i, C_i),i=1,...,n\}$ be the training data of the seen classes, where $X_i \in \mathcal{X}^{D}$ is the $D$-dimensional visual features (e.g., CNN features extracted from images) for the $i$-th image and $C_i \in \mathcal{C}^K$ is its $K$-dimensional class-level semantic features (e.g., class attribute embedding). Because one class usually corresponds to many image examples, we aim at finding a one-to-many class-to-instance feature generator. Specifically, we try to learn a mapping $g: \mathcal{C}^{K} \times \mathcal{Z}^{d} \rightarrow \mathcal{X}^{D}$ that seeks to explain the visual features $X_i$ extracted from each image by its corresponding class-level features $C_i$ and a $d$-dimensional vector of latent factors $Z_i \in \mathcal{Z}^{d}$ that accounts for instance variations. We assume $Z_i$ is sampled from a Gaussian prior distribution $N(0, I_d)$, where $I_d$ stands for the $d$-dimensional identity matrix. Once the generator $g$ learns to generate image features from class features, it can also generate $\widetilde{X}$ from any unseen classes. Formally, the feature-to-feature mapping can be formulated by a conditional latent variable model as follows:
\vspace{-0.2em}
\begin{equation}
\begin{aligned}
&Z \sim \mathcal{N}(0, I_d),\\
&X = g_{\theta}(C, Z) + \epsilon, \epsilon \sim \mathcal{N}(0,\sigma^2 I_D),\\
\end{aligned} 
\label{eq:model1}
\end{equation}
where $\theta$ contains all the learning parameters in the mapping function $g$, and $\epsilon$ is a $D$-dimensional noise vector following Gaussian distribution. The mapping $g$ can be any non-linear mapping. In this paper we adopt the top-down MLP parameterization of $g$, which is also called the conditional generator network, to map the latent factors $Z$ along with the class features $C$ to the visual features $X$. Generally, the model is defined by a prior distribution of latent factors $Z \sim p(Z)$ and the conditional distribution to generate visual features given the class features and the latent factors, i.e., $[X|Z, C] \sim p_{\theta}(X| Z, C)$. Let $q_{\text{data}}(X|C)$ be the true distribution that generates the training visual features given their associated class features. The goal of learning this model is to minimize the Kullback-Leibler divergence $\text{KL}(q_{\text{data}}(X|C)\Vert p_{\theta}(X|C))$ over $\theta$.
\subsection{Learning by Alternating Back-Propagation}
\subsubsection{Maximum Likelihood Learning}

The maximum likelihood estimation (MLE) of our model $p_{\theta}(X|C)$ is equivalent to minimizing the Kullback-Leibler divergence $\text{KL}(q_{\text{data}}(X|C)\Vert p_{\theta}(X|C))$ over $\theta$. 
The complete data model is given by
\vspace{-0.4em}
\begin{equation}
\begin{aligned}
&\log p_{\theta}(X,Z |C) = \log[p_{\theta}(X|Z, C)p(Z)]\\
&= -\frac{1}{2\sigma^2} \Vert X-g_{\theta}(C,Z) \Vert ^2 -\frac{1}{2}  \Vert Z \Vert ^2 + {\rm const},
\end{aligned} 
\end{equation}
where the constant term is independent of $X$, $Z$ and $\theta$.
The observed-data model or the log-likelihood is obtained by integrating out the latent factors $Z$: 
\vspace{-0.4em}
\begin{equation}
\log p_{\theta}(X|C) = \log \int p_{\theta}(X|Z,C)p(Z)dZ .
\end{equation}

Suppose we observe the training data $S=\{(X_i, C_i),i=1,...,n\}$, and $[X|C] \sim p_{\theta}(X|C)$, the goal of MLE is to maximize the observed-data log-likelihood:
\vspace{-1em}
\begin{equation}
\begin{aligned}
L(\theta)& = \sum_{i=1}^{n}  \log p_{\theta}(X_i|C_i).\\
\end{aligned}
\label{eq:likelihood} 
\end{equation}

Without loss of generality, we consider one observed pair $(X_i, C_i)$ and omit subscript $i$ for brevity. To maximize the log-likelihood, we adopt gradient ascent strategy. The gradient of the log-likelihood with respect to $\theta$ can be calculated by the following equation: 
\begin{equation}
\begin{aligned}
\frac{\partial}{\partial \theta}\log p_{\theta}(X|C) &= \frac{1}{p_{\theta}(X|C)}\frac{\partial}{\partial \theta}p_{\theta}(X|C)\\
&=\mathbb{E}_{Z \sim p_{\theta}(Z|X,C)}\left[ \frac{\partial}{\partial \theta} \log p_{\theta}(X,Z|C)\right]. 
\end{aligned} 
\label{eq:gradient}
\end{equation}
\vspace{-2.1em}
\subsubsection{Inferential Back-Popagation}
Since the expectation with respect to the posterior distribution $p_{\theta}(Z|X,C)$ in Equation (\ref{eq:gradient}) is analytically intractable, we resort to the average of the MCMC samples to approximate the expectation.
Specifically, we employ Langevin dynamics which carries out sampling by iterating:
\begin{equation}
\begin{aligned} 
Z_{\tau+1} &= Z_{\tau} +   \frac{s^2}{2} \frac{\partial}{\partial Z} \log p_{\theta}(X,Z_{\tau}|C) + sU_\tau,\\
&= Z_{\tau} + sU_{\tau} + \\ &\frac{s^2}{2}\left[ \frac{1}{\sigma^2}(X-g_{\theta}(C,Z_{\tau})) \frac{\partial}{\partial Z}g_{\theta}(C,Z_{\tau})-Z_{\tau}\right],
\end{aligned} 
\label{eq:inference}
\end{equation}
where $\tau$ denotes the time step for Langevin dynamics, $U_\tau$ is the Gaussian white noise corresponding to the Brownian motion, which is added to prevent the chain from being trapped by local modes, and $s$ is the step size. 

Because of the high computational cost of MCMC, it is infeasible to generate independent samples from scratch in each learning iteration. In practice, the MCMC transition of latent factors $Z$ in the current iteration starts from the previous updated result of $Z$, which is obtained from the previous learning iteration. We initialize $Z$ with Gaussian white noise at the beginning. Persistent MCMC update with such a warm start scheme is effective and efficient enough to provide fair samples from the posterior distribution.

We infer the latent factors $Z_i$ for each observed pair $(X_i, C_i)$ by sampling a single copy of $Z_i$ from $p_{\theta}(Z_i|X_i,C_i)$ via running finite steps of Langevin dynamics starting from the current $Z_i$ (i.e., the warm start). This is a conditional explaining-away inference solving an inverse problem, which is that given the specific visual features of an image and its associated class features, how to obtain its corresponding latent factors. Unlike VAE, our model does not need to recruit an extra network for inference.    

\subsubsection{Learning Back-Popagation} Once $Z$ is inferred, we learn the model via stochastic gradient algorithm by updating $\theta$ based on the Monte Carlo approximation of the gradient of $L(\theta)$ in Equation~(\ref{eq:gradient}):
\vspace{-0.4em}
\begin{equation}
\begin{aligned} 
&\theta_{t+1} = \theta_{t} +\gamma_t \frac{\partial}{\partial \theta}L(\theta),
\end{aligned} 
\label{eq:learning1}
\end{equation}
where $t$ is the time step for the gradient algorithm, $\gamma_t$ is the learning rate, and
\begingroup\makeatletter\def\f@size{9.0}\check@mathfonts
\def\maketag@@@#1{\hbox{\m@th\large\normalfont#1}}%
\begin{equation}
\begin{aligned} 
\frac{\partial}{\partial \theta}L(\theta) &\approx \sum_{i=1}^n \frac{\partial}{\partial \theta} \log p_{\theta}(X_i,Z_i|C_i)  \\
&= -\sum_{i=1}^n \frac{\partial}{\partial \theta}  \frac{1}{2 \sigma^2} \Vert X_i-g_{\theta}(C_i,Z_i) \Vert ^2   \\
& =\sum_{i=1}^n \frac{1}{\sigma^2}(X_i-g_{\theta}(C_i,Z_i)) \frac{\partial}{\partial \theta} g_{\theta}(C_i,Z_i) . 
\end{aligned} 
\label{eq:learning2}
\end{equation}
\endgroup
The Equation (\ref{eq:learning1}) corresponds to a non-linear regression problem, where it seeks to find the $\theta$ to predict visual features $X_i$ by its corresponding observed class features $C_i$ and the inferred latent factors $Z_i$.  

\subsubsection{Alternating Back-Propagation Algorithm}
The key to compute Equation (\ref{eq:inference}) is to calculate $\partial g_{\theta}(C,Z)/ \partial Z$, while the key to compute Equation (\ref{eq:learning2}) is to calculate $\partial g_{\theta}(C,Z)/ \partial \theta$. Both of them can be efficiently computed by back-propagation. Algorithm \ref{alg:abp} describes the details of the learning and sampling algorithm.  

\begin{algorithm}
	\caption{Alternating back-propagation procedure for learning feature-to-feature translator.}
	\label{code:3}
	\begin{algorithmic}[1]
		\REQUIRE training samples $\{(X_i, C_i), i=1,...,n\}$, the maximal number of loops $N_{step}$, %the batch size $m$, 
		the number of Langevin steps $l$, the learning rate $\gamma_t$.
		\ENSURE the learned parameters $\theta$, the inferred latent factors $\{Z_i, i =  1, ..., n\}$. 
		\item[]
		\STATE Initialize $\theta$ and $Z_i$, for $i =  1, ..., {n}$. 
		\FOR{t = 1,..., $N_{step}$}
		\STATE {\bf Inferential back-propagation}: For each $i$, run $l$ steps of Langevin dynamics to sample $Z_i \sim p_{\theta}(Z_i|X_i, C_i)$ with warm start, i.e., starting from the current $Z_i$, each step 
		follows equation (\ref{eq:inference}). 
		\STATE {\bf Learning back-propagation}: Update $\theta \leftarrow \theta + \gamma_t {L}'(\theta) $,  where ${L}'(\theta)$ is computed according to equation (\ref{eq:learning2}), with learning rate $\gamma_t$. 
		\ENDFOR
	\end{algorithmic}
	\label{alg:abp}
\end{algorithm}

\subsection{Comparison with Variational Inference}
 
Our learning algorithm presented in Algorithm \ref{alg:abp} seeks to minimize $\text{KL}(q_{\text{data}}(X|C) \Vert p_{\theta}(X|C))$, while variational auto-encoder (VAE) changes the objective to  
\begingroup\makeatletter\def\f@size{11.0}\check@mathfonts
\def\maketag@@@#1{\hbox{\m@th\large\normalfont#1}}%
\begin{equation*}
\begin{aligned} 
\min_{\theta, \phi} \text{KL}(q_{\text{data}}(X|C)p_{\phi}(Z|X,C)\Vert p(Z|C)p_{\theta}(X|Z,C))
\end{aligned} 
\label{eq:VAE}
\end{equation*}
\endgroup
by utilizing an extra inference model $p_{\phi}(Z|X,C)$ with parameter $\phi$ to infer latent variable $Z$, where the inference model $p_{\phi}(Z|X,C)$ is an analytically tractable approximation to $p_{\theta}(Z|X,C)$. Compared to the maximum likelihood objective $\text{KL}(q_{\text{data}}(X|C)\Vert p_{\theta}(X|C))$, which is the KL-divergence between the marginal distributions of $X$ conditioned on $C$, while the VAE objective is the KL-divergence between the joint distributions of $(Z,X)$ conditioned on $C$ (i.e., an upper bound of the maximum likelihood objective.) as shown below:
\begingroup\makeatletter\def\f@size{9.0}\check@mathfonts
\def\maketag@@@#1{\hbox{\m@th\large\normalfont#1}}%
\begin{equation*}
\begin{aligned} 
&\text{KL}(q_{\text{data}}(X|C)p_{\phi}(Z|X,C)\Vert p_{\theta}(Z, X|C)) = \\
&\text{KL}(q_{\text{data}}(X|C) \Vert p_{\theta}(X|C)) + \text{KL}( p_{\phi}(Z|X,C) \Vert p_{\theta}(Z|X,C) ).
\end{aligned} 
\label{eq:VAE}
\end{equation*}
\endgroup
The accuracy of variational inference depends on the accuracy of the inference model $p_{\phi}(Z|X,C)$ as an approximation of the true posterior distribution $p_{\theta}(Z|X,C)$. That is, when $\text{KL}( p_{\phi}(Z|X,C) \Vert p_{\theta}(Z|X,C))=0$, the variational inference is equivalent to the maximum likelihood solution. Therefore, our learning algorithm is more natural, straightforward, accurate, and computationally efficient than VAE.  
\section{Zero-Shot Learning}
\subsection{Classification in Zero-Shot Learning}

The feature-to-feature translator that maps $[C, Z] \rightarrow X$ can be considered as an explicit implementation of the local linear embedding \cite{roweis2000nonlinear}, where $[C, Z]$ is the embedding of $X$, with disentanglement of class features $C$ and non-class features $Z$. Given any unseen class $C^{u} \in \mathcal{C}^K$, we can generate arbitrarily many visual features for the unseen class by first sampling $Z$ from $\mathcal{N}(0,I_d)$ and then mapping $(C^{u}, Z)$ into $X$ via the learned feature-to-feature translator $\widetilde{X} = g_{\theta}(C^{u}, Z)+\epsilon$. With generated data of unseen classes, the labels of testing examples from unseen classes can be predicted via any conventional supervised classifiers. Here we employ a KNN classifier ($K$=20) for ZSL for its simplicity and a softmax classifier for GZSL as suggested in~\cite{xian2018feature}.

\subsection{Learning from Incomplete Visual Features}
The inferential back-propagation step performs an explaining-away inference, where the latent factors compete with each other to explain the observed visual features. This is very useful in the scenario where the training visual features are incomplete. In this case, our model is still able to learn from incomplete visual features via alternating back-propagation algorithm. The latent factors can still be obtained by explaining the incomplete observed visual features, and the model parameters can still be updated as before.   Taking the features in~\cite{zhu2018generative} as an example, Zhu \emph{et al} train a part detector to localize small semantic parts of birds such as heads and tails, and concatenate the extracted features of parts as the visual representation of the object for ZSL. The part features are inaccessible in two cases: 
(a) the semantic parts don't appear in the images, for example, no tails can be observed when birds are in front view; (b) the detector fails to discover the parts. In these cases, zeros are assigned to the corresponding bins of feature vector for the missing parts. We can easily adapt our ABP algorithm to the above situation by changing the computation of $\Vert X-g_{\theta}(C,Z) \Vert ^2$ to $\Vert M  \circ(X-g_{\theta}(C,Z)) \Vert ^2 $, where $M$ is the given binary indicator matrix with the same size of $X$, with 1 indicating ``visible'' and 0 indicating ``missing'', and sign $\circ$ denotes element-wise matrix multiplication operation. The indicator matrices $M_i$ vary for different visual features $X_i$. Note that GAN or VAE-based methods can hardly handle learning from incomplete visual features.

\section{Experiments}
\subsection{Experiment Settings}
\textbf{Datasets}. We evaluate the proposed framework for generative ZSL on four widely used ZSL benchmark datasets and compare it with a number of state-of-the-art baselines. The datasets include: (1) Caltech-UCSD Birds-200-2011 (CUB)~\cite{WahCUB_200_2011}, (2) Animal with Attributes (AwA1)~\cite{lampert2014attribute}, (3) Animal with Attributes 2 (AwA2)~\cite{xian2018zero}, (4) SUN attribute (SUN)~\cite{patterson2012sun}. 
CUB is a fine-grained dataset of bird species with 312 class-level attributes. AwA1 is a coarse-grained dataset including 50 classes of animals with 85 attributes. Since the original images in AwA1 are not publicly available due to the copyright license issue, Xian \emph{et al}~\cite{xian2018zero} create a new dataset AwA2 by collecting new images for each class in AwA1 while keeping the attribute annotations the same as AwA1. SUN contains 717 types of scenes with 102 attributes. We follow the train/test split settings in~\cite{xian2018zero}, which ensures that no unseen classes are included in the ImageNet dataset where the visual feature extractor is pretrained, to avoid violating the setting of ZSL.
Besides, we also conduct experiments on a large-scale dataset, i.e., ImageNet-21K~\cite{imagenet_cvpr09}. In this challenging dataset, no attribute annotations are available. We use the word embedding of the class names as the semantic represention of the classes~\cite{changpinyo2016synthesized}.

\textbf{Implementation Details}. Our translator is implemented by a multilayer perceptron with a single hidden layer of 4,096 nodes. LeakyReLU and ReLU are used as nonlinear activation functions on the hidden layer and the output layer respectively.  The dimension of latent factors $z$ is set to be 10. We fix
$\sigma$ = 0.3, $l$ = 10 and $s$ = 0.3. Our model is trained with a batch size of 64 and the Adam optimizer with a learning
rate of $10^{-3}$, $\beta_1$ = 0.9, and $\beta_2$ = 0.999. The number of epochs is 50.
Our model is implemented using Pytorch framework~\cite{paszke2017automatic} and trained on one NVIDIA TITAN Xp GPU. Our
code is publicly available online\footnote{\url{https://github.com/EthanZhu90/ZSL_ABP}} .

\subsection{Zero-Shot Learning}
As for zero-shot learning setting, we follow the evaluation protocol used in~\cite{xian2018zero}, where results are measured by average per-class top-1 accuracy. We compare with 15 state-of-the-art methods, including 11 non-generative ones and 4 generative ones which are separately shown in Table~\ref{table:zsl}.
For GAZSL, FGZSL, and MCGZSL, we get results by running their codes. We reimplement the VZSL by ourselves since no code is publicly available. The results of other methods are published in~\cite{xian2018zero}. From Table~\ref{table:zsl}, we can observe that: (i)  the generative methods have an overall performance superior to the non-generative ones (ii) our proposed model consistently outperforms previous state-of-the-art methods and shows great superiorities on AwA1 and AwA2, where the improvements are up to $3.1\%$. This validates that our model trained by alternating back-propagation is significantly beneficial to ZSL tasks. 

\begin{table}[t]
	\renewcommand{\arraystretch}{1.0}
	\begin{center}
		\scalebox{1.0}{
			\begin{tabular}{|c|l|cccc|}
				%\toprule
				\hline 
				&Method	 & CUB & AwA1 & AwA2 & SUN \\\hline 
				\multirow{10}{*}{$\mathcal{x}$}&DAP~\cite{lampert2014attribute}    & 40.0  & 44.1  & 46.1 & 39.9\\
				&CMT~\cite{socher2013zero}    & 34.6  & 39.5 & 37.9 & 39.9\\
				&LATEM~\cite{xian2016latent}  & 49.3  & 55.1 & 55.8 & 55.3\\
				&ALE~\cite{akata2016label}     & 54.9  & 59.9 & 62.5 &58.1\\
				&DEVISE~\cite{frome2013devise} & 52.0  & 54.2 & 59.7 &56.5\\
				&SJE~\cite{akata2015evaluation}   & 53.9  & 65.6 & 61.9 &53.7\\
				&ESZSL~\cite{romera2015embarrassingly}   & 53.9  & 58.2 & 58.6 & 54.5\\
				&SYNC~\cite{changpinyo2016synthesized}   & 55.6  & 54.0 &46.6 &56.3 \\ 
				&SAE~\cite{kodirov2017semantic}   & 33.3  & 53.0 & 54,1 &40.3 \\ 
				&DEM~\cite{zhang2017learning}     & 51.7  & 65.7 & 66.5 & 60.8 \\
				&GFZSL~\cite{verma2017simple} & 49.3 & 68.3 &63.8 &60.6\\\hline
				\multirow{5}{*}{$\mathcal{y}$} &VZSL~\cite{wang2017zero}     & 56.3  &\underline{67.1}  & 66.8 &59.0\\
				&GAZSL~\cite{zhu2018generative}  & 55.8  & 63.7  & 64.2 &\underline{60.1}\\
				&FGZSL~\cite{xian2018feature}  & 57.7  & 65.6  & 66.9 &58.6\\
				&MCGZSL~\cite{felix2018multi} & \underline{58.4}  &66.8 & \underline{67.3} &60.0 \\
				&Ours           & \textbf{58.5}  &\textbf{69.3}  & \textbf{70.4} &\textbf{61.5} \\
				\hline 
			\end{tabular}
		}
	\end{center}
	\vspace{-1em}
	\caption{Performance comparison for zero-shot learning on CUB, AwA1, AwA2 and SUN datasets. The performance is measured by average per-class top-1 accuracy (\%). $\mathcal{y}$ and $\mathcal{x}$ indicate generative and non-generative methods, respectively. The best and the second best results are marked in bold and underlined respectively.}
	\label{table:zsl}
	\vspace{-1em}
\end{table}

\subsection{Generalized Zero-Shot Learning}
In the generalized zero-shot learning setting, a test image is classified to the union of  seen and unseen classes.
This setting is more practical and difficult as it removes the assumption that test images only come from unseen classes. Following the protocol proposed by~\cite{xian2018zero}, we compute the harmonic mean of accuracies on seen and unseen classes: $H = \frac{2 \cdot A_{\mathcal{S}} \cdot  A_{\mathcal{U}}}{A_{\mathcal{S}} + A_{\mathcal{U}}}$, 
where $A_{\mathcal{S}}$ and $A_{\mathcal{U}}$ denote the accuracies of classifying images from seen classes and those from unseen classes respectively. We evaluate our method on four datasets and show the performance comparison with 13 state-of-the-art methods in Table~\ref{table:gzsl}. 
The experimental results show that non-generative methods achieving high performance on seen classes perform badly on unseen classes, indicating that those methods are biased in favor of seen classes. In contrast,  those generative methods can mitigate the bias and perform well on both unseen and seen classes. 
Compared with other generative ZSL methods, our method obtains much higher accuracies on unseen classes and comparable accuracies on seen classes. It improves the state-of-the-art performances by notable margins on most datasets (e.g., $4.0\%$ on AwA2 in terms of harmonic mean), demonstrating the capability for generalized zero-shot learning.

\begin{table*}[t]
	\renewcommand{\arraystretch}{1.0}
	\begin{center}
		\scalebox{1.0}{
			\begin{tabular}{|c|l|rrr|rrr|rrr|rrr|}
				\hline 
				&& \multicolumn{3}{c|}{CUB} & \multicolumn{3}{c|}{AwA1} & \multicolumn{3}{c|}{AwA2} &  \multicolumn{3}{c|}{SUN} \\
				&\multicolumn{1}{c}{Method}  & \multicolumn{1}{|c}{$A_{\mathcal{U}}$} & $A_{\mathcal{S}}$ & $H$ & \multicolumn{1}{c}{$A_{\mathcal{U}}$} & $A_{\mathcal{S}}$ & $H$ & \multicolumn{1}{c}{$A_{\mathcal{U}}$} & $A_{\mathcal{S}}$ & $H$ & \multicolumn{1}{c}{$A_{\mathcal{U}}$} & $A_{\mathcal{S}}$ & $H$\\ \hline 
				%\midrule
				\multirow{8}{*}{$\mathcal{x}$} &DAP \cite{lampert2014attribute}  & 1.7 & \textbf{67.9} & 3.3 & 0.0 &\textbf{88.7} & 0.0 
				& 0.0 & 84.7 & 0.0 & 4.2 & 25.1 & 7.2\\
				&DEVISE \cite{frome2013devise}  & 23.8 & 53.0 & 32.8 & 13.4 & 68.7 &22.4 
				& 17.1 & 74.7 & 27.8 & 16.9 & 27.4 & 20.9\\
				&CMT \cite{socher2013zero}  &7.2 &49.8 &12.6   &0.9	&\underline{87.6} &1.8 
				&0.5	&\textbf{90.0}	&1.0 &8.1	&21.8 &11.8\\
				&SJE \cite{akata2015evaluation}  & 23.5 & 59.2 & 33.6 & 11.3 & 74.6 &19.6 
				& 8.0 & 73.9 & 14.4 & 14.7 & 30.5 & 19.8\\
				&LATEM \cite{xian2016latent}  & 15.2 & 57.3 & 24.0 & 7.3 & 71.7 &13.3 
				& 11.5 & 77.3 & 20.0 & 14.7 & 28.8 & 19.5\\
				&ESZSL \cite{romera2015embarrassingly}   & 12.6 & \underline{63.8} & 21.0 & 6.6 & 75.6 &12.1 
				& 5.9 & 77.8 & 11.0 &11.0 & 27.9 & 15.8\\
				&ALE \cite{akata2016label}   & 23.7 & 62.8 & 34.4 & 16.8 & 76.1 &27.5 
				& 14.0 & 81.8 & 23.9 & 21.8 & 33.1 & 26.3\\
				&SAE \cite{kodirov2017semantic}  & 7.8 & 54.0 & 13.6 & 1.8 & 77.1 &3.5 
				& 1.1 & 82.2 & 2.2 & 8.8 & 18.0 & 11.8\\
				&DEM \cite{zhang2017learning}  &19.6	&57.9 &29.2  &32.8	&84.7 &47.3 
				&30.5	&\underline{86.4}	&45.1 &20.5	&34.3 &25.6\\
\hline
				\multirow{6}{*}{$\mathcal{y}$}
				&VZSL \cite{wang2017zero}  &44.9 &54.1 &49.1 &53.4 &68.3 &59.9 
				&51.7 &67.2 &58.4 &43.5 &34.9  &38.7	\\
				&GAZSL \cite{zhu2018generative}   &26.5	&57.4	&36.2	&32.8	&84.7	&47.3	&\textbf{59.9} &68.3 &53.4	&21.7	&34.5	&26.7\\
				&FGZSL \cite{kodirov2017semantic}   &\underline{45.9}	&54.6	&49.9	&53.1	&68.0	&59.6	&50.2	&67.5	&57.5	&40.2	&\underline{36.4}	&38.2\\
				&MCGZSL~\cite{felix2018multi}  &45.7  &61.0 &\textbf{52.3}	&\underline{56.9}&64.0&\underline{60.2}	&51.9 &67.2 &\underline{58.6}	&\textbf{49.4} &33.6 &\underline{40.0}\\
				&Ours   &\textbf{47.0}	&54.8	&\underline{50.6}	&\textbf{57.3}	&67.1	&\textbf{61.8} &\underline{55.3}	&72.6	&\textbf{62.6}	&\underline{45.3}	&\textbf{36.8}	&\textbf{40.6}\\\hline
			\end{tabular}
		}
	\end{center}
	\vspace{-1em}
	\caption{Performance comparison for generalized zero-shot learning.  $\mathcal{y}$ and $\mathcal{x}$ indicate generative and non-generative model-based methods respectively. The best and the second best results are marked in bold and underlined respectively.}
	\label{table:gzsl}

\end{table*}

\subsection{Large-Scale Experiments}
\label{sec:large}
We also evaluate the performance of our model on the large-scale ImageNet-21K~\cite{imagenet_cvpr09} dataset. The dataset contains a total of 14 million images from more than 21K classes. The relation among classes follows the WordNet~\cite{miller1995wordnet} hierarchy. Following the same protocol in ~\cite{xian2018zero, changpinyo2016synthesized}, we keep a specific subset of 1K classes for training, and use either all the remaining classes or a subset of it for testing. Specifically, the subsets for testing are determined according to the hierarchical distance from the training classes, or their population. For example, \textit{2Hop} contains 1,509 unseen classes that are within two tree hops of the seen 1K classes based on the class hierarchy, while \textit{3Hop} increases the number of unseen classes to 7,678 by extending the range to three tree hops. \textit{M500}, \textit{M1K} and \textit{M5K} contain 500, 1K, and 5K most populated classes, while \textit{L500}, \textit{L1K} and \textit{L5K} contain 500, 1K, and 5K least populated classes respectively.

\begin{figure}[t]
	\centering
	\begin{subfigure}[b]{0.235\textwidth}
		\includegraphics[width=1\columnwidth]{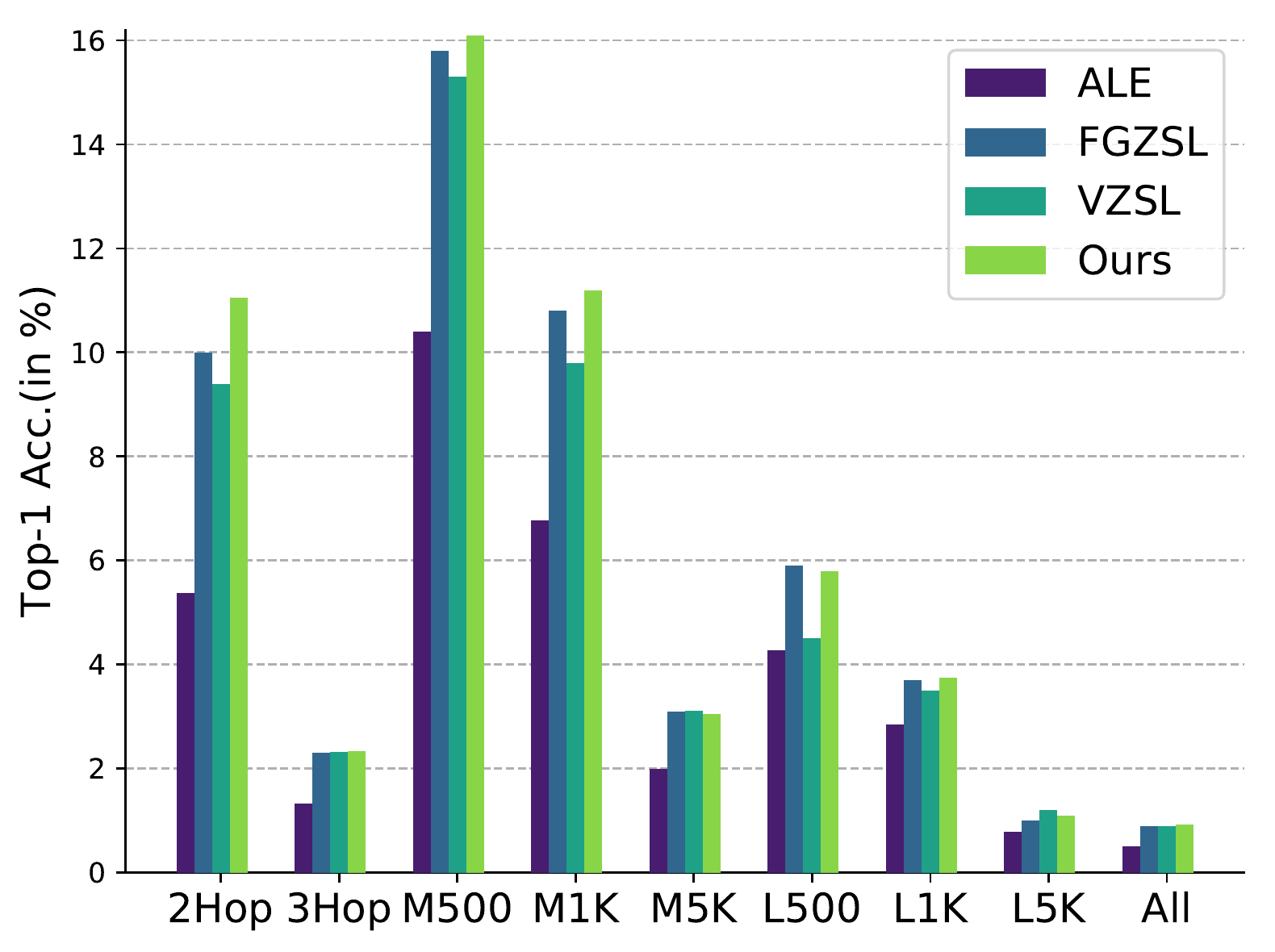}	
		\caption{ZSL}
		\label{fig:cub}
	\end{subfigure}
	%\hspace{-2em}
	\begin{subfigure}[b]{0.235\textwidth}
		\includegraphics[width=1\columnwidth]{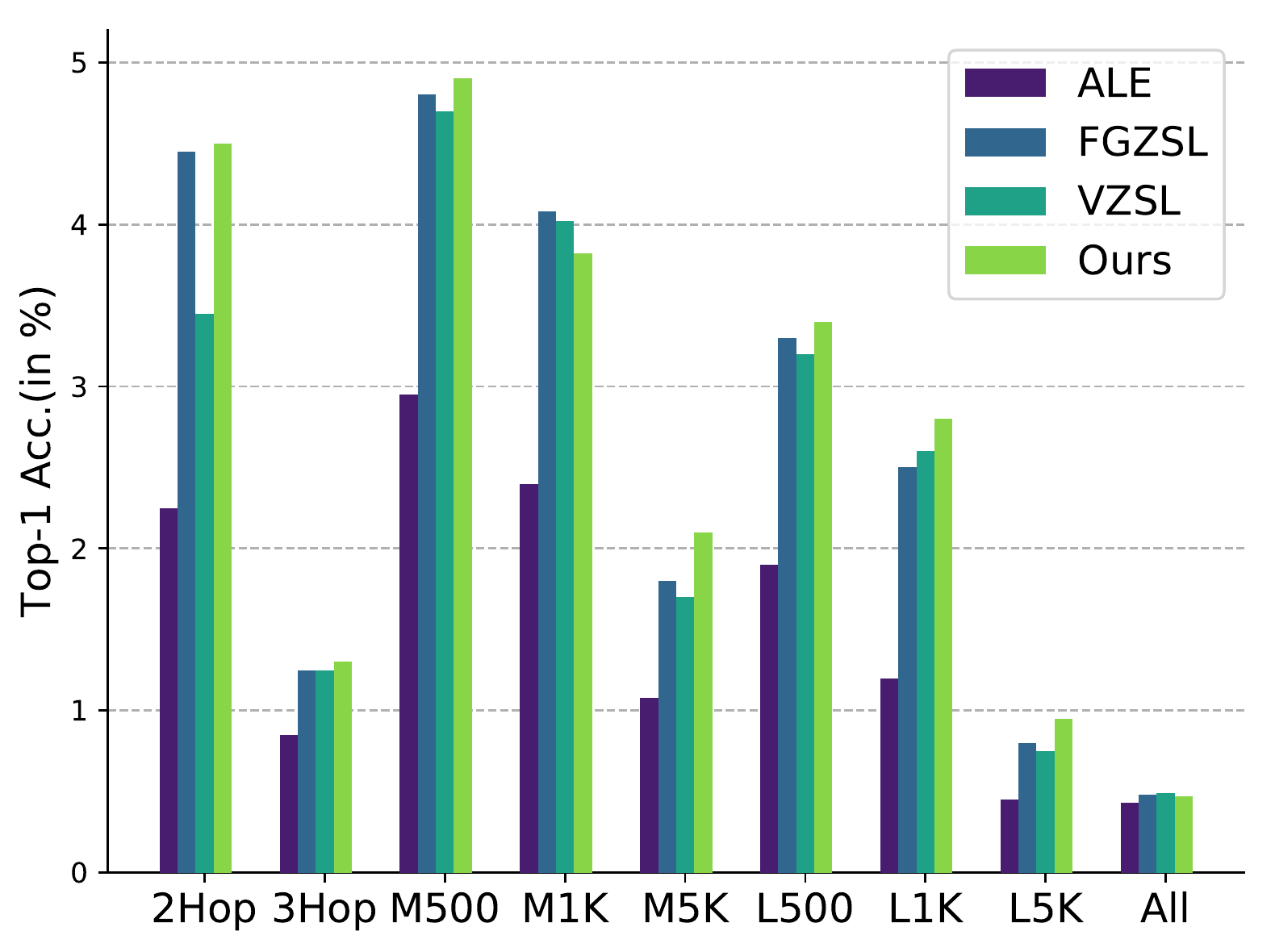}	
		\caption{GZSL}
		\label{fig:awa}
	\end{subfigure}
	%\hspace{-2em}

	\caption{ZSL and GZSL results on ImageNet. For GZSL, $A_{\mathcal{U}}$ is reported.}
	\label{fig:imagenet}
\end{figure}

We compare our method with three baseline methods, which include a visual-semantic embedding-based method, i.e., ALE, and two generative model-based methods, i.e., VZSL and FGZSL. The results are shown in Figure~\ref{fig:imagenet}. A notable observation is that all methods perform much better in \textit{2Hop} subset than in \textit{3Hop} subset. Two vital factors accounting for the observation are: in \textit{3Hop} subset, (i) the unseen classes are semantically less related to the training classes, making it difficult to transfer knowledge, (ii) a dramatic increase in the number of unseen classes (from $1,509$ to $7,678$) makes the classification even harder.  Generally, it is evident that generative methods have superior performances. Among the three generative methods, our proposed method achieves the best accuracies in most cases, demonstrating that it is very competitive in this realistic and challenging task.

\subsection{Comparison with GAN \& VAE-based Methods}
To thoroughly evaluate the performance of our model, we perform extensive ablation experiments. We study the comparison of our method with GAN and VAE-based methods (i.e., FGZSL and VZSL) in terms of \textbf{(i)} the speed of the convergence,\textbf{ (ii)} the number of model parameters, \textbf{ (iii)} ZSL performance when generating different numbers of features for unseen classes,\textbf{ (iv)} ZSL performance under different numbers of seen classes for training.

\begin{figure}[t]
	\centering
	\begin{subfigure}[b]{0.235\textwidth}
		\includegraphics[width=1\columnwidth]{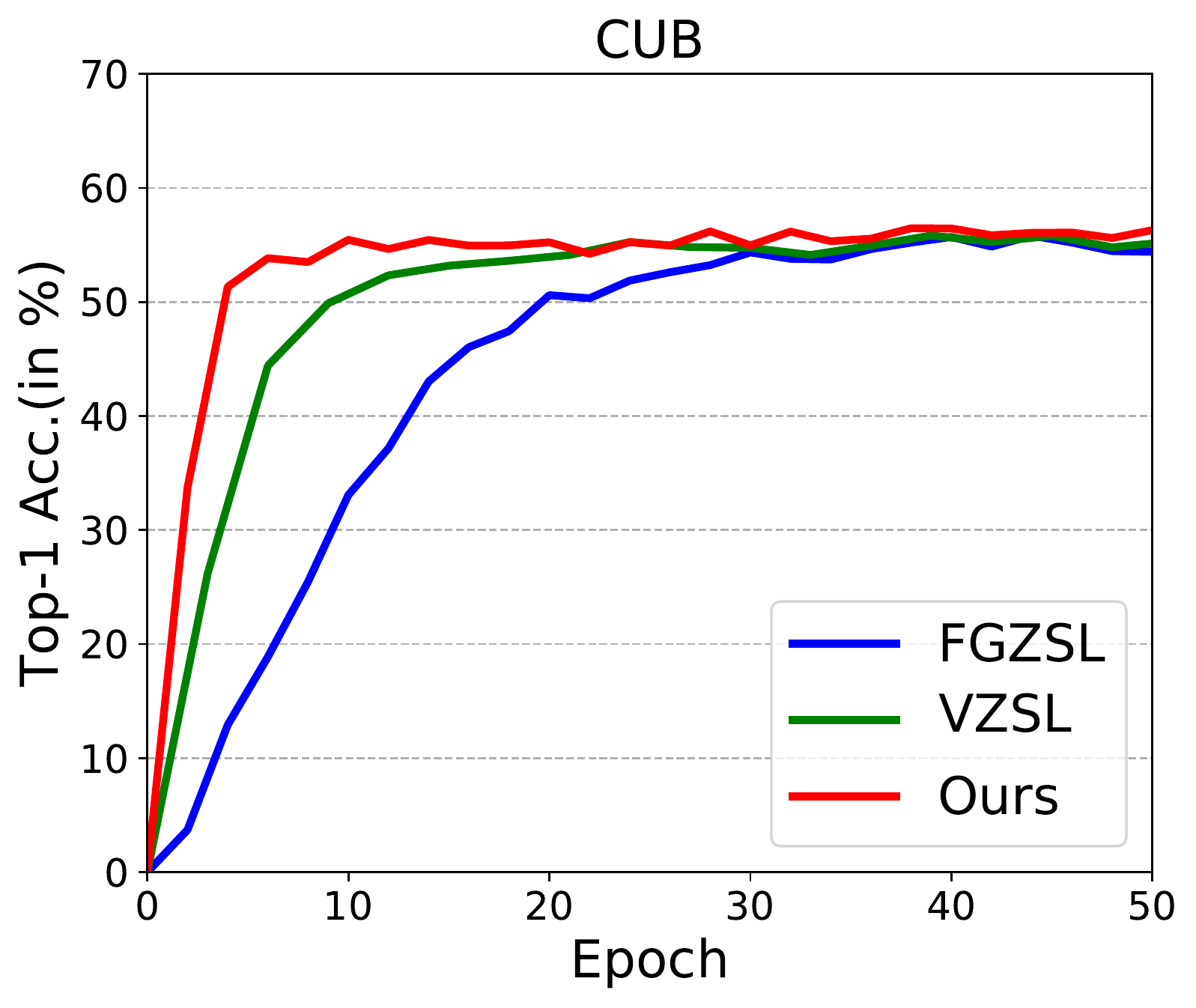}	
		\label{fig:cub}
	\end{subfigure}
	%\hspace{-2em}
	\begin{subfigure}[b]{0.235\textwidth}
		\includegraphics[width=0.99\columnwidth]{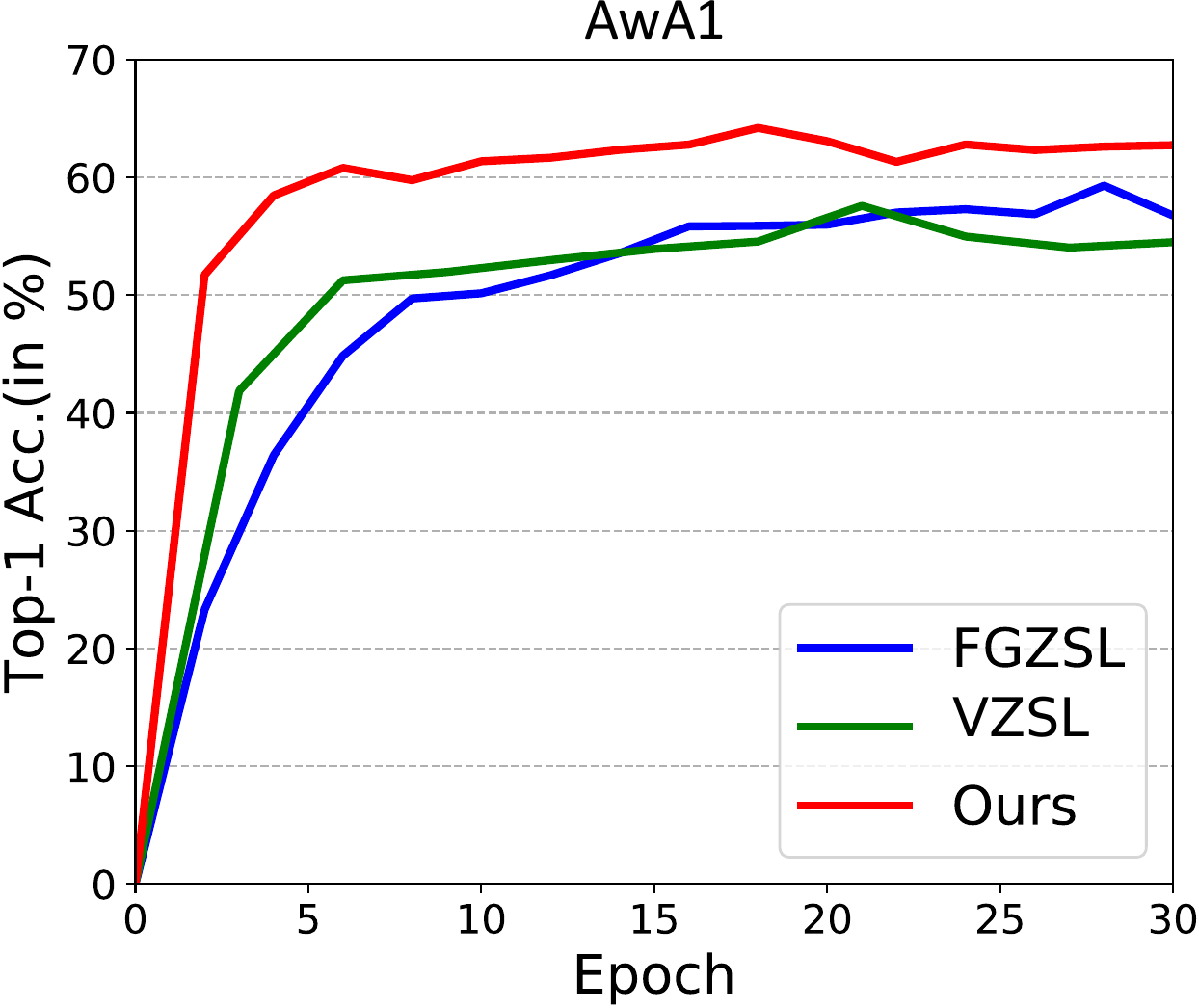}	
		\label{fig:awa}
	\end{subfigure}
	%\hspace{-2em}
	\vspace{-2.0em}
	\caption{Convergence comparison: top-1 accuracies in the validation set over different numbers of training epochs.}
	\label{fig:curve}
	\vspace{-1.0em}
\end{figure}
\textbf{\emph{Comparison (i)}} To investigate the advantage of our proposed method, we display the convergence curves of different models. We reserve a subset of seen classes as the validation set, and the remaining seen classes are used for training. 
Considering that various loss functions used in different methods are not comparable, we instead use the average per-class top-1 accuracy of the validation set as the convergence indicator. Figure~\ref{fig:curve} shows classification accuracies over different numbers of training epochs on CUB and AwA1 datasets. The convergence trends on both datasets are similar. Among these three methods, ours converges fastest, while the GAN-based method is the slowest one. 

\begin{table}[h]
	\begin{center}
				\scalebox{0.9}{
		\begin{tabular}{|l |c  |c | }
			\hline
		    Dataset	&\# of Parameters  &\# of Mult-Adds\\\hline
			FGZSL~\cite{xian2018feature}  & 20.62M & 41.23M \\
			VZSL~\cite{wang2017zero}   & 21.90M & 43.78M \\
			Ours   & 9.71M & 19.42M \\\hline
		\end{tabular}
	}
	\end{center}
	\vspace{-1em}
	\caption{Comparison of the number of parameters and the computational cost among three generative model-based ZSL methods that are applied to CUB dataset.}
	\label{table:param}

\end{table}

\textbf{\emph{Comparison (ii)}} Another advantage of our proposed model is the small number of parameters, or equivalently, the low computational load. We compare our method with FGZSL and VZSL for the CUB dataset in terms of  the number of parameters and the number of element-wise multiplication and addition operations in Table~\ref{table:param}. Note that these numbers might get changed when the methods are applied to other datasets, because different datasets may have various dimensions of semantic and visual features. Due to the incorporation of auxiliary networks in VAE and GAN-based methods (i.e., the encoder in VAE and the discriminator in GAN), these models require more parameters and computations. 
Our method only contains one single conditional generator, therefore its parameter size and computational cost are only half of those in GAN and VAE-based frameworks. 

\begin{figure}[t]
	\centering
	\begin{subfigure}[b]{0.235\textwidth}
		\includegraphics[width=1\columnwidth]{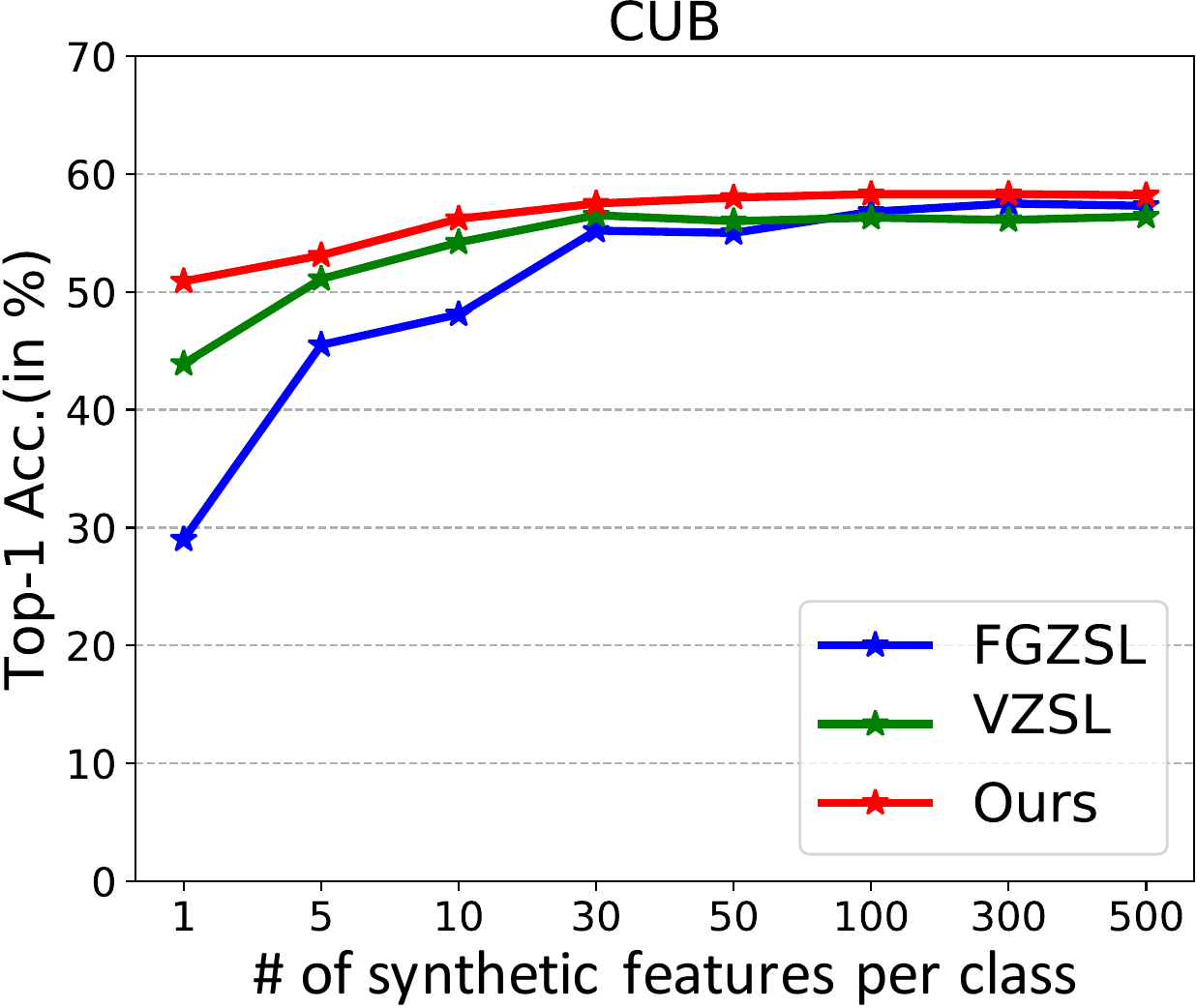}	
		\label{fig:cub}
	\end{subfigure}
	%\hspace{-2em}
	\begin{subfigure}[b]{0.235\textwidth}
		\includegraphics[width=0.99\columnwidth]{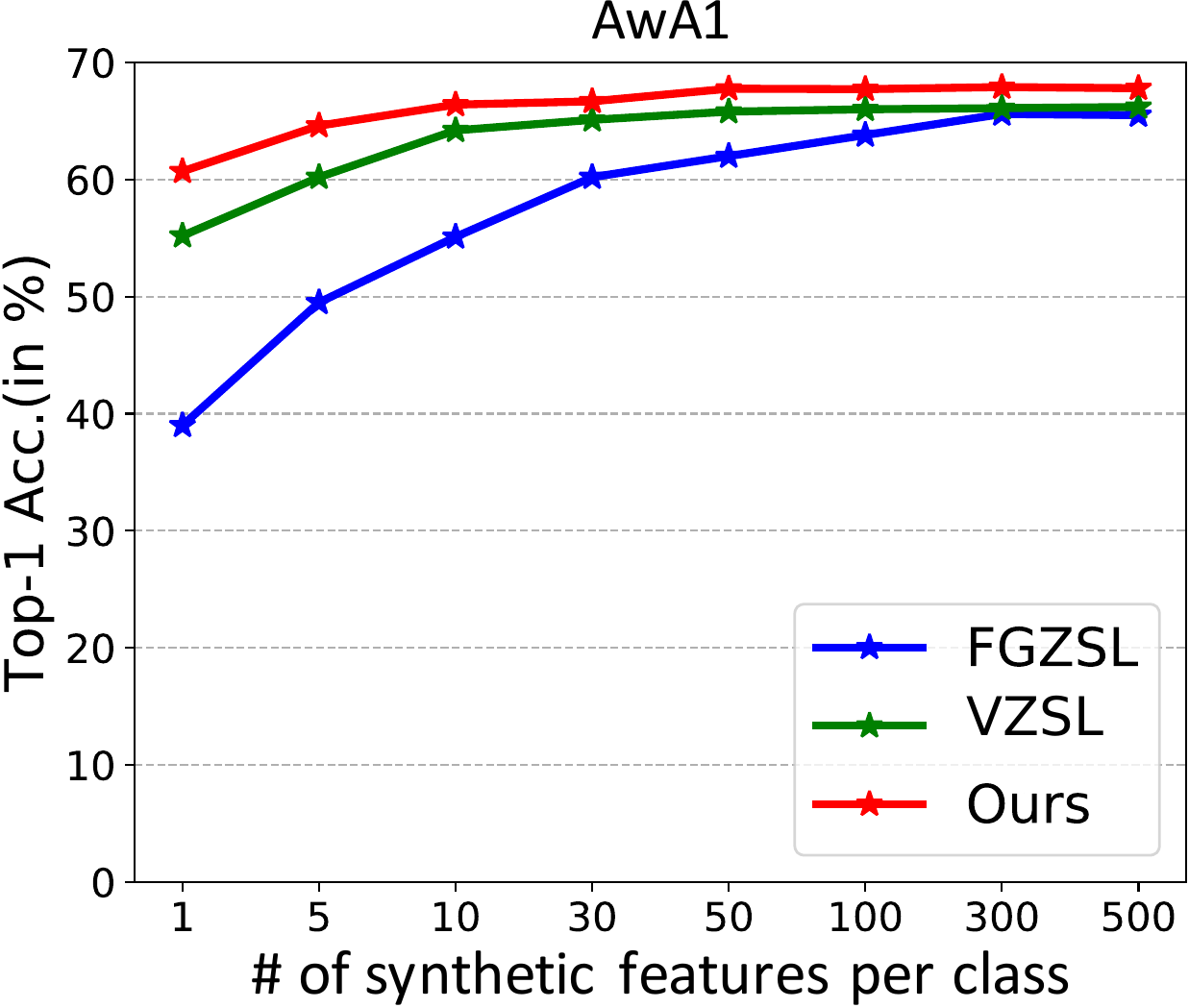}	
		\label{fig:awa}
	\end{subfigure}
	%\hspace{-2em}
	\vspace{-2.0em}
	\caption{ Comparison of top-1 per-class accuracies of unseen classes with different numbers of synthetic features per class.}
	\label{fig:pseudo}
		\vspace{-1.5em}
\end{figure}

\vspace{0.5em}
\textbf{\emph{Comparison (iii)}} Figure~\ref{fig:pseudo} shows the accuracies of three methods with different numbers of synthetic features for each unseen class. The same curve trends appear on both CUB and AwA1 datasets. The results are better with larger number of synthetic samples. When the number increases to 300, all the performances are similar and stable, indicating that each of the models is saturated as no improvement appears with more samples. Compared with FGZSL, our model performs much better with fewer samples (e.g., $50.1\%$ v.s. $29.1\%$ on CUB and $60.1\%$ v.s. $38.9\%$ on AwA1 with only 1 sample). The VZSL model performs slightly worse than ours. 
%\todo{add more explaination}

\begin{figure}[t]
	\centering
	%\hspace{-2em}
	\includegraphics[width=0.6\columnwidth]{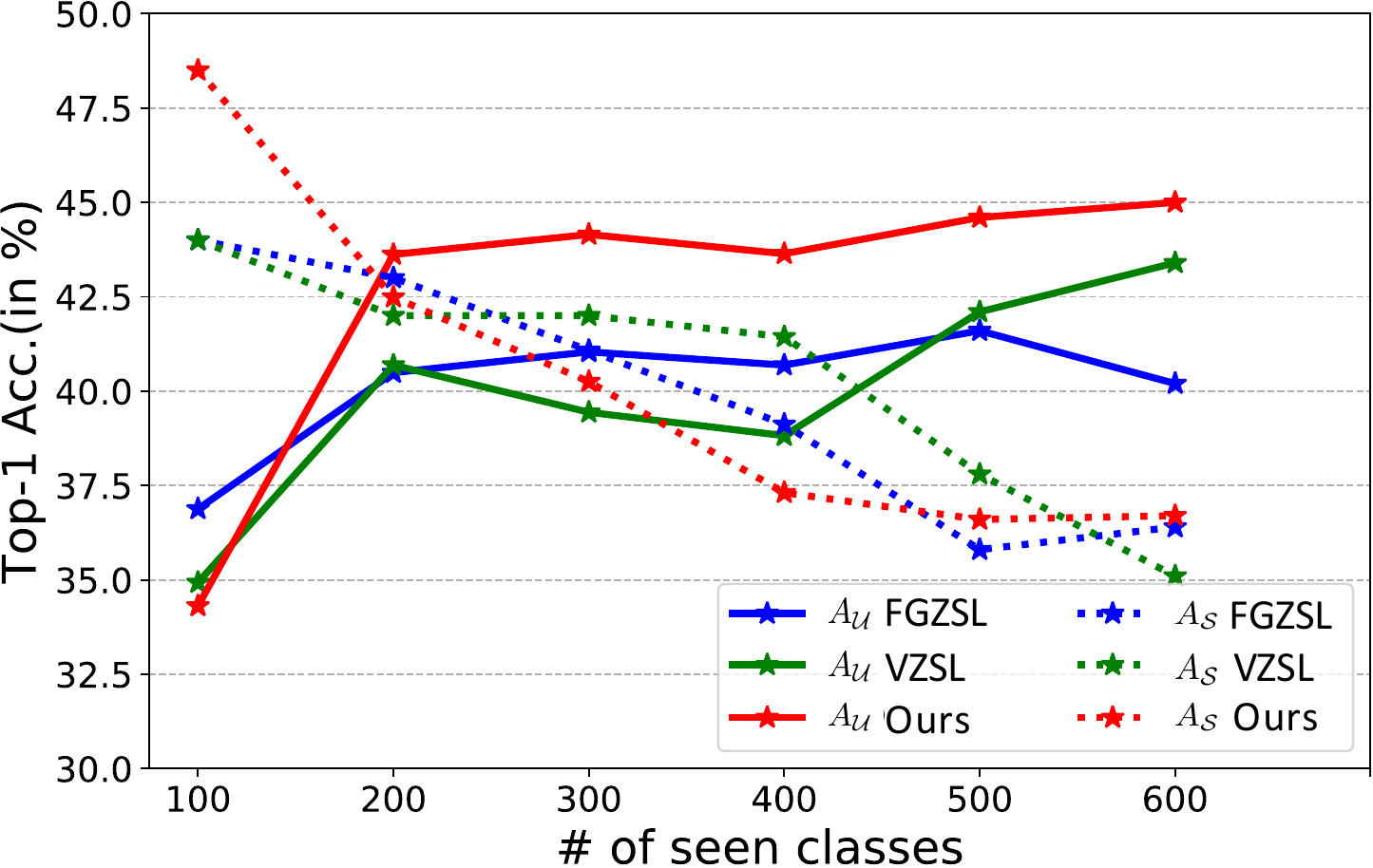}	
	\caption{Comparison of top-1 per-class accuracies of unseen classes among different methods with different numbers of seen classes in training.}
	\label{fig:class}
\end{figure}

\textbf{\emph{Comparison (iv)}} As pointed out in~\cite{song2018transductive}, the number of unseen classes usually dramatically surpasses that of seen classes in the real world. However, most ZSL benchmark datasets are far away from this situation. For instance, only 72 out of 717 classes on the SUN dataset are specified as unseen classes. This motivates us to investigate how models perform with different numbers of seen classes for training. 
We conduct experiments on the SUN dataset as it contains more classes than other datasets. We keep the unseen classes the same and randomly sample different numbers of seen classes for training. To make the experiments closer to the real world situation, we report the performance  in the generalized ZSL setting, as shown in Figure~\ref{fig:class}. As the number of seen classes increases, the accuracies of seen classes of all methods consistently decrease, indicating the increasing difficulty in discriminating seen class examples. From another perspective, more seen classes can provide more knowledge to associate the visual and semantic features, resulting in the improvement in the performance on unseen classes. Compared with other generative methods, our model achieves the best $A_{\mathcal{U}}$ while keeping decent $A_{\mathcal{S}}$.

\begin{comment}
\begin{figure}[h]
	\centering
	\includegraphics[width=0.9\columnwidth]{figures/tsne-crop.pdf}	
	\vspace{-1.0em}
	\caption{ t-SNE visulization for AwA1. (a) Origianl CNN features (b) Generated features by our model.}
	\label{fig:tsne}
	\vspace{-0.5em}
\end{figure}
\end{comment}

\subsection{Learning from Incomplete Visual Features}
Zhu \emph{et al} \cite{Elhosein_2017_CVPR, zhu2018generative} propose to use concatenated part features for ZSL, where those part features are either from groundtruth part annotations or extracted by a learned part detector, denoted as GTA and DET features respectively. The missing ratio in GTA is $9.3\%$, while the ratio decreases to $4.0\%$ in DET due to the high recall of their part detector. We first evaluate the performance of our model and GAZSL~\cite{zhu2018generative} with GTA features and DET features. As shown in Table~\ref{table:inc}, our method outperforms GAZSL by $2.6\%$ and $2.5\%$ using GTA and DET features respectively. To further analyze the power of our model in dealing with incomplete visual features, we increase the missing ratio of the DET features by randomly masking some valid feature values. As the missing ratio increases, the performances of both methods drop due to the reason that less and less useful information can be used. However, our method can still achieve a decent performance of $51.6\%$ in the most challenging situation where the missing ratio reaches $90\%$, while the accuracy of GAZSL is only $37.7\%$.  This confirms the claimed power of our model in learning from incomplete visual features for ZSL.

\begin{table}[t]
	\renewcommand{\arraystretch}{1.0}
	\begin{center}
		\scalebox{0.85}{
		\begin{tabular}{|lcc|cccc|}
			%\cline{4-7}	
		\hline
			& & &\multicolumn{4}{|c|}{DET$^*$}\\
			Method 
			& GTA &DET  &\multicolumn{1}{ |c }{$30\%$} &\multicolumn{1}{ c }{$50\%$} &\multicolumn{1}{ c}{$70\%$} &\multicolumn{1}{ c| }{$90\%$}\\ \hline
			GAZSL~\cite{zhu2018generative}
			& 74.1 & 72.7 &68.5 &63.7 &55.6 &37.7\\
			Ours    & 76.7 & 75.2  &72.9 &71.3 &64.8 &51.6\\
			\hline
		\end{tabular}
		}
	\end{center}
		\vspace{-1.2em}
	\caption{Zero-shot learning performance (top-1 accuracy $\%$) of the models trained on incomplete visual features with different missing ratios.}
	\label{table:inc}
		\vspace{-1.2em}
\end{table}

\section{Conclusion}
We propose a feature-to-feature translator learned by an alternating back-propagation algorithm as a general-purpose solution to zero-shot learning. Unlike other generative models, such as GAN and VAE, our method is simple yet effective, and does not rely on any assisting networks for training. The alternating back-propagation algorithm iterates the inferential back-propagation for inferring the instance-level latent factors and the learning back-propagation for updating the model parameters. We present a solution to learning from incomplete visual features for ZSL. We show that our framework outperforms the existing generative ZSL methods.

\textbf{Acknowledgment.} This work is partially supported by NSFUSA award 1409683. We thank Prof. Ying Nian Wu for helpful discussions.

{\small
\bibliographystyle{ieee}
\bibliography{egbib}
}

\end{document}